\documentclass{article}
\usepackage{graphicx}
\usepackage{latexsym}

\usepackage{amssymb}
\usepackage{amsmath}
\usepackage{algorithm}
\usepackage{algorithmic}
\usepackage{todonotes}
\usepackage{pgfplots}
\usetikzlibrary{patterns}
\usepackage{marvosym}
\usepackage{paralist}
\usepackage{tabularx}
\usepackage{url}

\usepackage{xspace}

\newcommand{\nop}[1]{}

\usepackage{scalefnt}
\usepackage{xspace}
\usepackage{relsize} 
\newcommand{\sysfont}{\textit}

\newcommand{\dlv}{\sysfont{DLV}\xspace}
\newcommand{\dlvdue}{\sysfont{DLV\smaller[0.8]{2}}\xspace}
\newcommand{\clingo}{\mbox{\sysfont{clingo}}\xspace}
\newcommand{\idlv}{{\mbox{\small $I$-\dlv}}\xspace}

\newif\ifrevisionmode
\revisionmodefalse

\ifrevisionmode
        \newcommand{\system}{\sysfont{System}\xspace}
\else
        \newcommand{\system}{\idlv}
\fi

\newcommand\derives{\ensuremath{\scalebox{0.6}[1.0]{\(:\!\!- \)}}}
\newcommand{\weak}{\ensuremath{:\sim}\xspace}
\newcommand{\Or}{\ensuremath{\ |\ }\xspace}
\newcommand{\p}{\ensuremath{{P}}\xspace}
\newcommand{\naf}{\ensuremath{\mathtt{not}}\xspace}

\newcommand{\cit}[1]{~\cite{#1}}
\newcommand{\re}[1]{~\ref{#1}}

\newcommand\quo[1]{{``{#1}''}}

\newif\ifrevisionmode
\revisionmodefalse

\ifrevisionmode
    
    \newcommand{\remove}[1] { {\color{red}\sout{#1}} }

    \newcommand{\TODO}[1]{{\color{red}{TODO: #1}}}
    \newcommand{\oldstuff}[1]{{\color{gray}FROM OLD PAPER: #1}}

\else

    \newcommand{\remove}[1]{}

    \newcommand{\TODO}[1]{}
    \newcommand{\oldstuff}[1]{{}}

\fi

\newcommand{\aggr}[1]{\ensuremath{\mathrm{\# #1}}}

\newenvironment{idlvsrsize}{\scriptsize\selectfont}{\par}
\newenvironment{monospace}{\fontfamily{qcr}\selectfont}{\par}

\usepackage{listings}
\lstdefinestyle{asp-style}{
	language=Prolog,
	keywordstyle=\linespread{1.1}\bf,
	basicstyle=\linespread{1.1}\em,
        columns=fullflexible,
	breaklines=true,
	mathescape=true,
	numbers=none,
	xleftmargin=0em,
        aboveskip=4pt,
        belowskip=4pt,
        morekeywords = {not,count,sum,@sum,min,max,div},
}
\lstdefinestyle{asp-style-inline}{
	language=Prolog,
	keywordstyle=\linespread{1.1}\bf,
	basicstyle=\linespread{1.1}\em,
	breaklines=true,
	mathescape,
	numbers=none,
	xleftmargin=0em,
        aboveskip=4pt,
        belowskip=4pt
}

\lstdefinestyle{python3}{
	language=Python,
	basicstyle=\linespread{1}\ttfamily,
	breaklines=true,
	mathescape,
	numbers=none,
	xleftmargin=0em,
        aboveskip=5pt,
        belowskip=5pt
}

\definecolor{dark-green}{rgb}{0,0.5,0}

\newcommand{\noninteger}{{non-integer}\xspace}
\newcommand{\nonintegers}{{non-integers}\xspace}
\newcommand{\rational}{{rational}\xspace}
\newcommand{\rationals}{{rationals}\xspace}

\newcommand{\float}{{floating-point}\xspace}

\title{Extending Answer Set Programming with Rational Numbers}

\author{Francesco Pacenza$^{0000-0001-6632-3492}$ and \\
Jessica Zangari$^{0000-0002-6418-7711}$}

\date{Department of Mathematics and Computer Science, \\ University of Calabria, Rende, Italy}

\begin{document}
\maketitle

\begin{abstract}
Answer Set Programming (ASP) is a widely used declarative programming paradigm that has shown great potential in solving complex computational problems. 
However, the inability to natively support \noninteger arithmetic has been highlighted as a major drawback in real world applications. 
This feature is crucial to accurately model and manage real world data and information as emerged in various contexts, such as the smooth movement of video game characters, the 3D movement of mechanical arms and data streamed by sensors.
Nevertheless, extending ASP in this direction, without affecting its declarative nature and its well-defined semantics, poses non-trivial challenges; thus, no ASP system is able to natively reason with \noninteger domains.
Indeed, the widespread floating-point arithmetic is not applicable to the ASP case, as reproducibility of results cannot be guaranteed and the semantics of an ASP program would not to be uniquely and declaratively determined, regardless of the employed machine or solver. 
To overcome such limitations and in the realm of pure ASP, this paper proposes an extension of ASP in which \nonintegers are approximated to \rational numbers, fully granting reproducibility and declarativity.  
We provide a well-defined semantics for the ASP-Core-2 standard extended with \rational numbers and an implementation thereof.
We hope this work could serve as a stepping stone towards a more expressive and versatile ASP language that can handle a broader range of real world problems.

\end{abstract}

\section{Introduction}

In recent years, artificial intelligence (AI) has become an integral part of our daily lives, influencing various aspects of society, from personal interactions to business operations. As AI technologies continue to advance, there are ongoing efforts towards explainability. 
One of the key challenges in achieving transparency is the inherent complexity of AI algorithms. 
Many AI systems operate as \quo{black boxes}, making it challenging for users and even developers to comprehend the decision-making processes. 

In this light, Answer Set Programming (ASP)~\cite{DBLP:journals/cacm/BrewkaET11,DBLP:conf/rweb/EiterIK09,gelf-lifs-1991,DBLP:books/sp/Lifschitz19} is a declarative symbolic language, promoted as a good candidate to address the explainability issue\cit{DBLP:conf/ijcai/EiterG23}.
ASP is a rule-based programming paradigm, in which logic rules are used to explicitly describe problems. Rules drive the inference process, making explainable the resulting solutions, thus facilitating a clear understanding of the decision-making process.
The roots of ASP are grounded in the area of knowledge representation and reasoning and in particular, in logic programming and non-monotonic reasoning.
Applications of ASP can be found in many real world contexts, such as industry\cit{DBLP:journals/ki/FalknerFSTT18,DBLP:conf/padl/TakeuchiBTS23}, robotics\cit{DBLP:conf/kr/AngilicaABIP23,DBLP:journals/ki/ErdemP18}, planning\cit{DBLP:journals/tplp/Bogatarkan020}, scheduling\cit{DBLP:conf/padl/Yli-JyraRJ23}, IoT\cit{DBLP:journals/ijimai/CostantiniGL21}, stream reasoning\cit{DBLP:journals/ai/BeckDE18}, surgery\cit{DBLP:journals/ml/MeliSF21}, diagnosis\cit{DBLP:journals/apin/WotawaK22}, psychology\cit{DBLP:journals/tplp/Inclezan15}, video-games\cit{DBLP:conf/padl/AngilicaIPZ23} and more\cit{DBLP:journals/aim/ErdemGL16}.

In the realm of explainability, novel applications are emerging from the increasing effort of the scientific community towards the integration of symbolic and statistical AI approaches~\cite{DBLP:conf/sefm/Maruyama20,DBLP:conf/kr/IshayY023}.
However, diverse attempts\cit{DBLP:conf/ijcai/Bourneuf18,DBLP:journals/aim/ErdemGL16,DBLP:journals/ki/FalknerFSTT18,DBLP:journals/ml/MeliSF21} have been emphasizing the absence in ASP of an essential requirement, i.e., the support for non-integer numbers.
Non-integer arithmetic is already supported by other logic programming languages, e.g., Constraint Logic Programming\cit{DBLP:journals/jlp/JaffarM94}, Prolog\cit{DBLP:journals/tplp/WielemakerSTL12}, Datalog\cit{DBLP:conf/ecoop/PacakE22}. 
Moreover, also hybrid approaches, merging the advances in distinct research areas such as Constraint Processing and Satisfiability Modulo Theories~\cite{a16040185,lierler_2023_translational,DBLP:journals/amai/MellarkodGZ08}, provide some support for \nonintegers. 
However, ASP-Core-2, the standard language~\cite{DBLP:journals/tplp/CalimeriFGIKKLM20} for ASP, adopted in the official competition series, does not feature \noninteger arithmetic, and no ASP-based system is currently able to natively handle numbers beyond integers. ASP developers are thus required to design ad-hoc solutions tailored to each specific application, when possible, or resort to approximations.

In fact, extending ASP in this direction, without affecting its declarative nature and its well-defined semantics, poses non-trivial challenges from both the theoretical and practical perspectives.
A natural solution could be to rely on the floating-point arithmetic for which technical standards are available, making it widely adopted not only in imperative programming but also in popular logic-based languages, such as Prolog\footnote{\url{https://www.swi-prolog.org/FAQ/floats.html}}.
However, the absence of some fundamental mathematical properties, like the associativity of addition and multiplication, represents a critical issue in the ASP case, because it can lead to imprecise or incorrect results.
Moreover, reproducibility cannot be guaranteed regardless of the employed machine or implementation while, 
in contrast, the semantics of an ASP program should be uniquely and declaratively determined. 

In this regard, this work provide the following contributions.
\begin{inparaitem}
\item[$i)$] 
    We define a new version of the ASP-Core-2 language including \rational numbers, formalizing its syntax and semantics. 
    Adopting \rational numbers is valuable under several aspects. 
    They possess the same properties of integers, while additionally providing closure with respect to all four fundamental mathematical operations.
    From a mathematical perspective rationals provide exact calculations. 
    In turn, from a computational point of view, like integers, if unlimited memory is available, rational numbers do not overflow and provide exact calculations in all circumstances.
    Lastly, they significantly broaden the range of application domains compared to the solely integers of ASP-Core-2.
\item[$ii)$] 
    We illustrate relevant aspects to take into account in order to implement the proposed language and to re-adapt an existing ASP system to support \rational terms.
\item[$iii)$] 
    In order to assess the practicality and viability of the introduced language, we provide a new version of the grounder \system\cit{DBLP:journals/ia/CalimeriFPZ17} supporting it. 
    \system is compatible with state-of-the-art ASP solvers and with any other solver adopting the de-facto standard \textit{lparse}\cit{DBLP:conf/lpnmr/SyrjanenN01} intermediate numeric format.
    Therefore, through the combination of \system with various solvers, the AI community gains access to multiple ASP systems able to handle numbers beyond integers.
    \system is available at \url{https://demacs-unical.github.io/I-DLV}.
\end{inparaitem}

The remainder of the paper is structured as follows.
Section\re{sec:rationals} recalls ASP and presents the conceived ASP-Core-2 version with \rational terms. Section\re{sec:motivations} discusses typically adopted solutions to bypass the lack of \noninteger numbers in ASP. 
Section\re{sec:implementation} presents the proposed implementation.
Section\re{sec:conclusions} concludes the work.

\section{ASP-Core-2 with Rational Terms}\label{sec:rationals}
Answer Set Programming is a purely declarative logic formalism rooted in logic programming and non-monotonic reasoning, falling in the area of Knowledge Representation and Reasoning.
The basic construct of ASP is a rule, that has the form $Head \leftarrow Body$ where the $Body$ is a logic conjunction in which negation may appear, and $Head$ can be either an atomic formula or a logic disjunction. 
A rule is interpreted according to common sense principles: roughly, its intuitive semantics corresponds to an implication. 
More precisely, the semantics of ASP is called \textit{answer set semantics}. 
In ASP a problem is modeled via a logic program composed by a collection of rules. 
An ASP system is in charge of determining its solutions by computing its \textit{answer sets}, which correspond one-to-one to a solution of the modeled problem. 
In order to determine the answer sets of a given program, the majority of ASP systems relies on an {\em instantiation} (or {\em grounding}) phase followed by a {\em solving} phase.
In the former step, a {\em grounder} module produces a propositional program semantically equivalent to the input program; in the latter step, a {\em solver} module applies search techniques on this propositional program to determine the answer sets\cit{DBLP:conf/iclp/SilverthornLS12}.
Other approaches are based on {\em lazy grounding}\cit{DBLP:conf/aaai/BomansonJW19}, translation to propositional clauses\cit{DBLP:conf/birthday/JanhunenN11} or reduction to epistemic programs\cit{DBLP:conf/ecai/BesinHW23}.


Throughout the years a significant effort has been spent in order to extend the \quo{basic} language and ease knowledge representation tasks with ASP. 
The standard input language for ASP systems has been defined under the name of ASP-Core-2, the official language of the ASP Competition series~\cite{DBLP:journals/jair/GebserMR17}. 
It is worth to mention that ASP-Core-2 as well as mainstream ASP systems, supports only integers as numeric type and results of arithmetic operations are rounded as integers towards zero.


The remainder of this section presents the syntax and the semantics of proposed language extension, conceived on the basis of ASP-Core-2.
Then, we focus on technical aspects relevant for implementations.

\subsection{Syntax}\label{sec:asp_syntax}
Let $\mathcal{I}$ be a set of \textit{identifiers}. 
An identifier is a non-empty string starting with some lowercase letter and containing only alphanumeric symbols and the symbol ``\_''.

A term is either a \textit{constant}, a \textit{variable}, an \textit{arithmetic term} or a \textit{functional term}. 
In particular, constants and variables can be considered as ``basic terms'', while arithmetic and functional terms are inductively defined as combinations of terms.
A \textit{constant} is either a \textit{rational}, a \textit{symbolic constant} if it is an identifier, a \textit{string constant} if it is a quoted string.
A \textit{rational} can be written in three forms: $(1)$ $p/q$ where $p$, $q\neq 0$ are integers representing the numerator and the denominator, respectively; $(2)$ $i$ where $i$ is an integer; $(3)$ $i.d_1...d_m$ where $i$ is an integer, $m>0$, each $d_j$ ($1\leq j\leq m$, $0\leq d_j\leq 9$) is called decimal digit.
Forms $(2)$ and $(3)$ are just short-hands for form $(1)$; indeed, in form $(2)$ the numerator is $i$ and the denominator is $1$; similarly, form $(3)$ corresponds to form $(1)$ with $(i.d_1...d_m*10^m)$ as numerator and $10^m$ as denominator.
A \textit{variable} is a non-empty string starting with some uppercase letter and containing only
alphanumeric symbols and the symbol ``\_''.
Furthermore, an \textit{anonymous variable} is a special form of variable, denoted by the symbol ``\_'' and is intended to indicate a \textit{fresh} variable that does not appear elsewhere in the context in which it is located.
An \textit{arithmetic term} has the form $-(t_1)$ or $(t_1 \diamond t_2)$ for terms $t_1$ and $t_2$, and $\diamond \in \{$\quo{$+$}$,$\quo{$-$}$,$\quo{$*$}$,$\quo{$/$}$\}$. 
Parentheses can optionally be omitted and standard operator precedences are applied.
A \textit{functional term} has the form $f(t_1,\dots,t_n)$, where $f$ is an identifier, called \textit{functor}, $t_1,\dots,t_n$ are terms and $n>0$.
A term is \textit{ground} (i.e., variable-free) if it does not contain any variable. 

Given an identifier $p$ and an integer $n$ with $n\geq0$, the expression $p/n$ represents a \textit{predicate}. 
$p$ is said \textit{predicate symbol} and $n$ represents the associated arity.
A \textit{predicate atom} has the form $p(t_1,\dots,t_n)$ where $n\geq0$, $p/n$ is a predicate and $t_1,\dots,t_n$ are terms; if $n=0$, parenthesis are omitted and the notation $p$ is used.
A \textit{classical atom} is either $-a$ or $a$ where $a$ is a predicate atom and $-$ denotes the \textit{strong negation} symbol.
A \textit{built-in atom} has the form $t_1\vartriangleright t_2$ where $t_1$, $t_2$ are terms and ${\vartriangleright} \in \{$\quo{$<$}$,$ \quo{$\leq$}$,$\quo{$=$}$,$\quo{$\neq$}$,$\quo{$>$}$,$\quo{$\geq$}$\}$.
A \textit{naf-literal} can either be a built-in atom or have form $a$ or $\naf\ a$ where $a$ is a classical atom, and $\naf$ is the \textit{negation as failure} symbol.
An \textit{aggregate element} is composed as: $t_1,\ldots,t_m : l_1,\ldots,l_n$, where $t_1,\ldots,t_m$ are terms $l_1,\ldots,l_n$ are naf-literals for $n\geq0$, $m\geq0$.
An \textit{aggregate atom} has the form:
$$\#aggr\{e_1; \ldots; e_n\} \vartriangleright t $$
where:
\begin{itemize}
\item $\#aggr \in \{$ \quo{$\aggr{count}$}$, $\quo{$\aggr{sum}$}$, $\quo{$\aggr{max}$}$,\allowbreak $\quo{$\aggr{min}$}$ \}$;
\item $e_1; \ldots; e_n$ are aggregate elements for $n\geq0$;
\item ${\vartriangleright} \in \{$\quo{$<$}$,$ \quo{$\leq$}$,$\quo{$=$}$,$\quo{$\neq$}$,$\quo{$>$}$,$\quo{$\geq$}$\}$;
\item $t$ is a term.
\end{itemize}
An \textit{aggregate literal} is either $a$ or $\naf\ a$ where $a$ is an aggregate atom.
In the following we will refer to classical, built-in and aggregate atoms simply as \textit{atoms}.
Similarly, we will indicate naf and aggregate literals as \textit{literals}. 
An atom is \textit{ground} if it does not contain any variable. 
A literal is ground if its atom is ground. 
A literal is \textit{negative} if the $\naf$ symbol is present, otherwise it is \textit{positive}.

A \textit{rule} $r$ has the following form:
$$a_1 \Or \ldots \Or a_n\ \derives\ b_1, \ldots, b_m.$$
where:
\begin{itemize}
\item $a_1, \ldots, a_n $ are classical atoms;
\item $b_1, \ldots, b_m $ are literals;
\item $n\geq0,m\geq0$.
\end{itemize}
The \textit{disjunction} $a_1\Or\ldots\Or a_n$ is the \textit{head} of $r$, while the \textit{conjunction} $b_1, \ldots, b_m$ is the \textit{body} of $r$. 
We denote by $H(r)$ the set $\{a_1 ,\ldots, a_n \}$ of head atoms, and by $B(r)$ the set $\{b_1, \ldots, b_m\}$ of body literals. 
$B^+(r)$ (respectively, $B^-(r)$) denotes the set of atoms occurring in positive (respectively, negative) literals in $B(r)$. 
A \textit{fact} is a rule $r$ with $B(r)=\emptyset$, $|H(r)|=1$ and $H(r)=\{a\}$ with $a$ ground.
A \textit{strong constraint} is a rule $r$ with $|H(r)|=\emptyset$.
A \textit{weak constraint} has the form: $\weak\ b_1, \ldots, b_m. \ [w@l,t_1,\ldots,t_n]$ where:
\begin{itemize}
\item $n\geq0$, $m\geq0$;
\item $b_1, \ldots, b_m$ are literals;
\item $w, l, t_1, \ldots, t_n$ are terms; $w$ and $l$ are referred to as \textit{weight} and \textit{level}, respectively; if $l=0$, the expression $@0$ can be omitted.
\end{itemize}
For a weak constraint $c$ we will indicate as \textit{weak specification}, denoted $W(c)$, the part within the square brackets.
A rule or weak constraint is \textit{ground} if no variable appears in it. 
A \textit{program} is a finite set of rules and weak constraints\footnote{We omit the concept of query whose semantics remains exactly as in ASP-Core-2; indeed, the presence of a query is transparent to the presence of \rational terms.}.
A program is \textit{ground} if all its rules and weak constraints are ground. 

\subsection{Semantics}\label{sec:asp_semantics}
The semantics of an ASP program is given by the set of its \textit{answer sets}, as reported in this subsection.

\medskip

\noindent Let \p be an ASP program.

\medskip \noindent {\bf Herbrand universe.} 
The \textit{Herbrand universe} of \p, $U_P$, is the set of all rational numbers in their standard form and ground terms constructible from constants and functors appearing in \p. 
We recall that a rational number is in the standard form if the numerator and denominator have no factors in common (except $1$) and the denominator is positive. 
In other words, for an integer $i$ its standard form is $\frac{i}{1}$, whereas for a rational number $\frac{p}{q}$ with $p$ integer, its standard form is $\frac{p'}{q'}$, where: $p'=p/gcd(p,q)$ and $q'=q/gcd(p,q)$ if $q$ is a non-zero positive integer; $p'=-p/gcd(p,q)$ and $q'=-q/gcd(p,q)$ if $q$ is a non-zero negative integer. Note that $gcd$ stands for \textit{greatest common divisor}.

\medskip \noindent {\bf Herbrand base.} 
The \textit{Herbrand base} of \p, $B_P$, is the set of all ground classical atoms obtainable by combining predicate names appearing in \p with terms from $U_P$ as arguments.

\medskip \noindent {\bf Global and local variables.}
For a literal $l$, let $var(l)$ be the set of variables appearing in $l$; if $l$ is ground $var(l)=\emptyset$. 
For a conjunction of literals $C$, $var(C)$ denotes the set of variables occurring in the literals in $C$; similarly, for a disjunction of atoms $D$, $var(D)$ denotes the set of variables in the atoms in $D$. 
Inductively, for a rule $r$, $var(r)=var(H(r))\cup var(B(r))$; for a weak constraint $c$, $var(c)=var(B(c))\cup var(W(c))$.
Given a rule or a constraint $r$, a variable is \textit{global} if it appears outside of an aggregate element in $r$ and we denote as $var_g(r)$ the set of global variables in $r$. 
Given an aggregate element $e$ in a rule or a weak constraint $r$, $var_l(e) = var(e) \setminus var(r)$ denotes the set of \textit{local} variables of $e$, i.e., the set of variables appearing only in $e$; intuitively, the set of \textit{global} variables of $e$ contains variables appearing in both $r$ and $e$, i.e., $var_g(e) = var(r) \cap var(e)$. 

\medskip \noindent {\bf Substitution.} 
Given a Herbrand universe $U_P$ of a program \p and a set of variables $V$, a \textit{substitution} is a total function $\sigma : V \mapsto U_P$ that maps each variable in $V$ to an element in $U_P$.
For some object $o$ occurring in \p (term, atom, literal, rule, etc.), we denote by $o\sigma$ the object obtained by replacing each occurrence of a variable $v \in var(o)$ by $\sigma(v)$ in $o$.

\medskip \noindent {\bf Global and local substitutions.} 
Given a rule or weak constraint $r$ in \p a substitution is \textit{global} if it involves variables in $var_g(r)$; for an aggregate element $e$ in $r$, a substitution is \textit{local} if it involves variables in $var_l(e)$.
We remark that for terms, classical atoms and naf-literals, a substitution is implicitly global, due to the absence of aggregate elements. 

\medskip \noindent {\bf Well-formed global and local substitutions.}\label{par:wellformed}
A global substitution $\sigma_g$ for a rule or weak constraint $r$ is well-formed if the arithmetic evaluation of any arithmetic subterm $-(t)$ or $(t\diamond u)$ appearing outside of aggregate elements in $r\sigma_g$ is well-defined.
Similarly, a local substitution for an aggregate element $e$, $\sigma_l$ is well-formed if the arithmetic evaluation of any arithmetic subterm $-(t)$ or $(t\diamond u)$ appearing in $e\sigma_l$ is well-defined.
In both cases, the arithmetic evaluation is performed in the standard way and results are reduced to the standard form.
More in detail, if $t$ and $u$ are rationals of form $p_t/q_t$ and $p_u/q_u$, respectively, the result is a rational term $v$ of form $p_v/q_v$ where:
\begin{itemize}
    \item in the $-(t)$ case:
    \begin{itemize}
        \item $p_v=-1*p_t$
        \item $q_v=q_t$
    \end{itemize}
    \item if $\diamond=\{$\quo{$+$}$\}$:
    \begin{itemize}
        \item $p_v=p'/gcd(p',q')$
        \item $q_v=q'/gcd(p',q')$
    \end{itemize}
    with $p'=(lcm(q_t,q_u)\ /\ q_t*p_t+lcm(q_t,q_u)\ /\ q_u*p_u)$ and $q'=lcm(q_t,q_u)$
    \item if $\diamond=\{$\quo{$-$}$\}$:
    \begin{itemize}
        \item $p_v=p'/gcd(p',q')$
        \item  $q_v=q'/gcd(p',q')$
    \end{itemize}
   with $p'=(lcm(q_t,q_u)\ /\ q_t*p_t-lcm(q_t,q_u)\ /\ q_u*p_u)$ and $q'=lcm(q_t,q_u)$
    \item if $\diamond=\{$\quo{$*$}$\}$:
        \begin{itemize}
            \item $p_v=(p_t*p_u)/gcd(p_t*p_u,q_t*q_u)$
            \item $q_v=(q_t*q_u)/gcd(p_t*p_u,q_t*q_u)$
        \end{itemize}
    \item if $\diamond=\{$\quo{$/$}$\}$:
        \begin{itemize}
            \item $p_v=(p_t*q_u)/gcd(p_t*q_u,q_t*p_u)$
            \item $q_v=(q_t*p_u)/gcd(p_t*q_u,q_t*p_u)$
        \end{itemize}
\end{itemize}
in which $lcm$ stands for \textit{least common multiple}.

\medskip \noindent {\bf Instantiation of aggregate elements.} 
The instantiation of a collection of aggregate elements $E$ is obtained by considering well-formed local substitutions for each aggregate element in $E$:
$$inst(E) = \bigcup_{e\in E} \{ e\sigma | \sigma \text{ is a well-formed local substitution for } e\}$$

\medskip \noindent {\bf Standardization of rationals.} 
Given the set $Q$ of \rational terms appearing in \p, a \textit{standardization} is a total function $\phi : Q \mapsto U_P$ mapping each \rational term $t \in Q$ to the element in $U_P$ corresponding to its standard form. Given an object $o$ occurring in \p (term, atom, literal, rule, etc.), we denote by $o\phi$ the object obtained by replacing each \rational term $t$ appearing in $o$ by $\phi(t)$ in $o$.

\medskip \noindent {\bf Ground instance.} 
A \textit{ground instance} of a rule or weak constraint $r \in \p$ is obtained in three steps: $(1)$, a standardization $\phi$ is applied to $r$ obtaining $r\phi$; $(2)$, a well-formed global substitution $\sigma_g$ for $r\phi$ is applied to $r\phi$ obtaining $r\phi\sigma_g$; $(3)$, for every aggregate atom $\#aggr E \vartriangleright t $ in $r\phi\sigma_g$, $E$ is replaced by $inst(E)$.

\medskip \noindent {\bf Arithmetic evaluation.} 
The arithmetic evaluation of a ground instance $g$ of a rule or a weak constraint is obtained by replacing any maximal arithmetic sub-term appearing in $g$ by its rational value in its standard form, which is calculated as shown in the paragraph related to well-formed global and local substitutions.

\medskip \noindent {\bf Instantiation of a program.} 
The \textit{ground instantiation} of a program \p, denoted by $grnd(P)$, is the set of arithmetically evaluated ground instances of rules and weak constraints in \p.

\medskip \noindent {\bf Interpretation.}
Once that a ground program is obtained, the truth values of atoms, literals, rules, constraints etc., are properly defined according to interpretations.
An \textit{(Herbrand) interpretation} $I$ for \p is a subset of $B_P$.

\medskip \noindent {\bf Total order.}
Literals can be either true or false w.r.t. an interpretation. 
To illustrate how their truth values are determined, as a preliminary step, we need to define a proper total order $\preceq$ on terms in $U_P$. 
In line with ASP-Core-2, we adopt the one reported next.
Let $t$ and $u$ be two arithmetically evaluated ground terms, then:
\begin{itemize}
    \item $t \preceq u$ for rationals $t$ of form $p_t/q_t$ and $u$ of form $p_u/q_u$ if $p_t * q_u \leq p_u * q_t$,
    \item $t \preceq u$ if $t$ is a rational and $u$ is a symbolic constant,
    \item $t \preceq u$ for symbolic constants $t$ and $u$ with $t$ lexicographically smaller or equal to $u$,
    \item $t \preceq u$ if $t$ is a symbolic constant and $u$ is a string constant,
    \item $t \preceq u$ for string constants $t$ and $u$ with $t$ lexicographically smaller or equal to $u$,
    \item $t \preceq u$ if $t$ is a string constant and $u$ is a functional term,
    \item $t \preceq u$ for functional terms $t=f(t_1,\ldots,t_n)$ and $u=g(u_1,\ldots,u_n)$ if either:
    \begin{itemize}
        \item $m<n$ or,
        \item $m=n$ and $g\npreceq f$ ($f$ is lexicographically smaller than $g$) or,
        \item $m=n$, $f\preceq g$ and, for any $1 \leq j \leq m$ s.t. $t_j \npreceq u_j$, there is some $1 \leq i < j$ s.t. $t_i \npreceq u_i$ (i.e., the tuple of terms of $t$ is smaller than or equal to the arguments of $u$).
    \end{itemize}
\end{itemize}
At this point, we are ready to properly define satisfaction of literals. 

\medskip \noindent {\bf Satisfaction of naf-literals.}
Let $I \subseteq B_P$ be an interpretation for \p.
The satisfaction of a built-in atom can be defined according to the total order $\preceq$, in the intuitive way, as they represent comparisons among terms; more in detail: 
\begin{itemize}
    \item $t\leq u$ is true w.r.t. $I$ if $t \preceq u$, false otherwise;
    \item $t\geq u$ is true w.r.t. $I$ if $u \preceq t$, false otherwise;
    \item $t < u$ is true w.r.t. $I$ if $t \preceq u$ and $u \npreceq t$, false otherwise;
    \item $t > u$ is true w.r.t. $I$ if $u \preceq t$ and $t \npreceq u$, false otherwise;
    \item $t = u$ is true w.r.t. $I$ if $t \preceq u$ and $u \preceq t$, false otherwise;
    \item $t \neq u$ is true w.r.t. $I$ if $t \npreceq u$ or $u \npreceq y$, false otherwise.
\end{itemize}
A classical atom $a \in B_P$ is true w.r.t. $I$ if $a \in I$; false w.r.t. $I$ otherwise. 
A positive naf-literal $a$ is true w.r.t. $I$ if $a$ is a classical or built-in atom that is true w.r.t. $I$; otherwise, $a$ is false w.r.t. $I$. 
A negative naf-literal $\naf\ a$ is true (or false) w.r.t. $I$ if $a$ is false (or true) w.r.t. $I$.

\medskip \noindent {\bf Satisfaction of aggregate literals.}
An aggregate function stands for a mapping from sets of tuples of terms to terms, $+\infty$ or $-\infty$. 
Each aggregate function maps a set $T$ of tuples of terms to a term, $+\infty$ or $-\infty$ as follows.
Let $T$ be a finite set of tuples of terms, then:
\begin{itemize}
    \item $\#count(T) = |T|$;
    \item $\#sum(T) = x'$ and $x = \sum_{(t_1,\ldots,t_m) \in T, t_1 \text{ is a rational}} t_1$ if $\{(t_1,\ldots,t_m)\in T | t_1 \text{is a non-zero rational}\}$ is finite and $x'$ is $x$ reduced to the standard form;
    \item $\#max(T) = max\{ t_1 | (t_1,\ldots,t_m) \in T\}$ if $T\neq \emptyset$; $\#max(T) = -\infty$ if $T=\emptyset$;
    \item $\#min(T) = min\{ t_1 | (t_1,\ldots,t_m) \in T\}$ if $T\neq \emptyset$; $\#min(T) = +\infty$ if $T=\emptyset$;
\end{itemize}
When $T$ is infinite, instead we have:
\begin{itemize}
    \item $\#count(T) = +\infty$;
    \item $\#sum(T) = 0$ if $\{(t_1,\ldots,t_m)\in T | t_1 \text{is a non-zero rational}\}$ is infinite;
    \item $\#max(T) = +\infty$;
    \item $\#min(T) = -\infty$.
\end{itemize}
We adopt the same convention of ASP-Core-2: $-\infty \preceq u$ and $u \preceq +\infty$ for every term $u \in U_P$.
Essentially, $\#count$ depends on the cardinality of the set of tuples of terms $T$, $\#sum$ is evaluated as the sum of rational terms in $T$ reduced in its standard form, while $\#max$ and $\#min$ functions strictly rely on the total order $\preceq$ on terms in $U_P$. 
Given an expression $\#aggr(T) \vartriangleright u$ s.t. $\#aggr \in \{$\quo{$\#count$}, \quo{$\#sum$}, \quo{$\#max$}, \quo{$\#min$}$\}$ and $u$ is a term, it is true (or false) according to the definition given for the satisfaction of built-in atoms, extended to the values $+\infty$ and $-\infty$ for $\#aggr(T)$.
We can now define the satisfaction of aggregate literals. 
Fixed an interpretation, some aggregate elements may not contribute to the semantics of an aggregate atom. 
Intuitively, an interpretation can filter out some aggregate elements according to their truth values w.r.t. the interpretation itself. 
More formally, the interpretation $I$ maps a collection $E$ of aggregate elements to the following set of tuples of terms:
\begin{equation*}
    \begin{split}
    eval(E,I) = &\{(t_1,\ldots,t_n) | \{t_1,\ldots,t_n : l_1,\ldots,l_m\} \text{ occurs in } E \text{ and } \\
    &\{l_1,\ldots,l_m\} \text{ are true w.r.t. } I\}
    \end{split}
\end{equation*}

Let $a= \#aggr E \vartriangleright t$ be an aggregate atom, $a$ is true (or false) w.r.t. $I$ if $\#aggr\{eval(E,I)\} \vartriangleright t$ is true (or false) w.r.t. $I$. 
A positive aggregate literal $a$ is true (or false) w.r.t. $I$ if $a$ is true (or false) w.r.t. $I$. 
A negative aggregate literal $\naf\ a$ is true (or false) w.r.t. $I$ if $a$ is false (or true) w.r.t. $I$.



\medskip
\noindent
{\bf Satisfaction of rules.}
Let $r$ be a rule in $grnd(P)$. The head of $r$ is \textit{true} w.r.t. $I$ if $H(r) \cap I \neq \emptyset$. The body of $r$ is \textit{true} w.r.t. $I$ if all body literals of  $r$ are true w.r.t. $I$ (i.e.,  $B^+(r) \subseteq I$ and  $B^-(r)\cap I = \emptyset$) and is \textit{false} w.r.t. $I$ otherwise. The rule $r$ is \textit{satisfied} (or \textit{true}) w.r.t. $I$ if its head is true w.r.t. $I$ or its body is false w.r.t. $I$.

\medskip
\noindent
{\bf Model.}
A \textit{model} for \p is an interpretation $M$ for \p such that every rule $r \in grnd(P)$ is true w.r.t. $M$.
A model $M$ for \p is \textit{minimal} if no model $N$ for \p exists such that $N$ is a proper subset of $M$. 
The set of all minimal models for \p is denoted by ${\rm MM}(P)$.

\medskip \noindent {\bf Reduct.}
Given the ground program $grnd(\p)$ and an interpretation $I$, the \textit{reduct} of $grnd(\p)$ w.r.t. $I$ is the subset $P^I$ of $grnd(\p)$, which is obtained from $grnd(\p)$ by deleting rules in which a body literal is false w.r.t. $I$.
Note that the above definition of reduct\cit{DBLP:journals/ai/FaberPL11} is equivalent to the Gelfond-Lifschitz transform for the definition of answer sets\cit{gelf-lifs-1991}.

\medskip \noindent {\bf Answer set.}
Let $I$ be an interpretation for \p. $I$ is an \textit{answer set} for \p if $I \in {\rm MM}({P^I})$. 
The set of all answer sets for \p is denoted by $AS(P)$.

\medskip \noindent {\bf Optimal answer sets.}
In case of weak constraints in \p, answer sets need to be further examined, and classified as
\textit{optimal} or not. Intuitively, strong constraints represent conditions that {\em must} be satisfied in every answer set, while {\em weak constraints} indicate conditions that {\em should} be satisfied; their semantics involves minimizing the number of violations, thus allowing to easily encode optimization problems.

Optimal answer sets of \p are selected among $AS(P)$, according to the following schema. 
Let $I$ be an interpretation, then:

\begin{equation*}
\begin{split}
weak(P,I) = &\{(w@l,t_1,\ldots,t_m) | \\ &\weak b_1,\ldots,b_n [w@l,t_1,\ldots,t_m]\text{ occurs in }grnd(P)\\
& \text{ and }b_1,\ldots,b_n\text{ are true w.r.t. }I
\}
\end{split}
\end{equation*}

\noindent For any rational $l$, we define $P^{I}_{l}$ as:
$$\sum_{(w@l,t_1,\ldots,t_m) \in weak(P,I),\ w \text{ is a rational }} w$$
if $\{(w@l,t_1,\ldots,t_m) \in weak(P,I)\ |\ w$ is a non-zero rational$\}$ is finite;
$0$ if $\{(w@l,t_1,\ldots,t_m) \in weak(P,I)\ |\ w$ is a non-zero rational$\}$ is infinite.
In other words, for each weak constraint in $grnd(P)$ satisfied by $I$, we sum the weights per level: these numbers represent a kind of penalty paid by $I$: the lower they are, the higher is the possibility for $I$, if it represents an answer set, to be optimal.

More formally, we define the notion of \textit{domination} among answer sets as follows. Given an answer set $A \in AS(P)$, it is said \textit{dominated} by another answer set $A'$ if there is some rational $l$ such that $P^{A'}_{l} < P^{A}_{l}$ and $P^{A'}_{l'} = P^{A}_{l'}$ for all rationals $l'> l$.
An answer set $A\in AS(P)$ is optimal if there is no $A'\in AS(P)$ such that $A$ is dominated by $A'$.

\subsection{Properties}\label{sec:properties}
With respect to ASP-Core-2, the proposed extensions enjoys the properties given in the following propositions.

\medskip
\noindent
\textbf{Proposition 1.} {\em The proposed semantics, exactly like ASP-Core-2, is based on a countably infinite set of numerals.}

\medskip
\noindent
\textbf{Proof.} Every rational number in the standard form can be uniquely associated with an integer, meaning there is a bijective relationship between the set of integers and the set of rationals in the standard form. Consequently, from the Cantor-Schroeder-Bernstein theorem,
it follows that the set of rationals in the standard form is countably infinite, exactly like the set of integers. 
Therefore, our proposal, in line with standard ASP with integers only as numeric type, relies on a countably infinite set of numerals.

\medskip
\noindent
\textbf{Proposition 2.} {\em If a program contains only rational numbers with denominator 1, i.e., integers, and only integer division is allowed, the proposed semantics coincides with the ASP-Core-2 semantics.}

\medskip
\noindent
\textbf{Proof.}
The set of integer numbers is closed under addition, subtraction and multiplication, but not under division.
Thus, the only way to obtain a rational number with a denominator different that $1$ is in the division case of the definition of well-formed global and local substitutions. In line with ASP-Core-2, we can set the numerator as the result of the integer division between $p_v$ and $q_v$ and the denominator to 1.

\subsection{Technical Aspects}\label{sec:technical}
We discuss below some aspects to take into account for the actual implementation of the proposed ASP-Core-2 extension.

\medskip \noindent {\bf Safety.} In ASP, the concept of safety has been introduced in order to limit possible substitutions of variables.
We inherit the safety restriction from ASP-Core-2~\cite{DBLP:journals/tplp/CalimeriFGIKKLM20}, as the herein proposed extension acts on only ground term types, extending the set of numerals from integers to rationals. 

\medskip \noindent {\bf Range and Modulus Operators.} The range \quo{$..$} and modulus \quo{$\backslash$} operators over  rationals would produce infinite results, making them unusable in practice. 
We thus limit these operators to integers and inherit their definitions from previous work~\cite{DBLP:conf/birthday/LifschitzLS20}. 
To exemplify such infinity issues, consider the rule: 
\begin{lstlisting}[style=asp-style]
a(X) :- X = 1..3.
\end{lstlisting}
in which a range operator is used to make $X$ vary between $1$ and $3$. 
Intuitively, if $X$ is grounded as an integer it could take as values $1$, $2$ and $3$, thus the grounding is finite;
if instead $X$ were required to range over rationals, the grounding would be infinite. 
Similarly, the same issue happens with the modulus operator. 

\medskip \noindent {\bf Undefined arithmetics.}
In line with ASP-Core-2, we require that programs should be invariant under undefined arithmetics, that is, their semantics is invariant regardless the handling of not well-formed substitutions.
For instance, let us consider the program $P_1$:
\begin{lstlisting}[style=asp-style]
a(0).
a(1/2).
c(Z) $\derives$ a(X), a(Y), Z = X / Y.
\end{lstlisting}
Intuitively, every substitution in which $Y\mapsto 0$ is not well-formed as it implies a division by 0 and $P_1$ results not invariant under undefined arithmetics.
By modifying the rule in $P_1$ as: 
\begin{lstlisting}[style=asp-style]
c(Z) $\derives$ a(X), a(Y), Z = X / Y, Y != 0.
\end{lstlisting}
$P_1$ would result invariant under undefined arithmetics as not well-formed substitutions are not applicable.

\medskip \noindent {\bf Number of decimal digits.}
As defined in Section\re{sec:asp_syntax}, a \rational term of form $(3)$, $i.d_1...d_m$, is transformed as a fraction with $(i.d_1...d_m*10^m)$ as numerator and $10^m$ as denominator. 
Clearly, in practical implementations, a fixed maximum number of digits, say $f$, can be kept in memory, and in order to guarantee reproducibility, systems should use the same $f$ and agree about how to round numbers when $f<m$.
While this implies that the semantics of the program might vary based on the value of $f$, this form is of great relevance in real world scenarios where data can come from external sources, e.g., from sensors.
It is worth to note that, if in the input program no rational of form $(3)$ appears, the semantics of such a program is always the same, no matter which is the value of $f$, the adopted reasoner or the underlying machine.
Moreover, for a fixed value of $f$ and an established rounding policy, the semantics of every program is independent from the adopted machine or reasoner.
In our implementation the default value of $f$ is fixed to $6$ and numbers with more than $f$ digits are rounded to $f$ decimal places, i.e., the $f$-th digit is rounded to the nearest.
In addition, the system allows to specify a different value for $f$ via a command line option.

\medskip \noindent {\bf Intermediate numeric format.}
The introduction of rational terms is transparent for the solving phase and compatibility with already available ASP solvers is guaranteed upon few updates in the common \textit{lparse}\cit{DBLP:conf/lpnmr/SyrjanenN01} intermediate numeric format used by grounders to pass the computed grounding to solvers.
These updates are needed because the numeric format has been conceived to work on integers only and concern the so-called weight rules that encode $\#sum$ aggregates, and minimize rules used for  weak constraints (see \cit{DBLP:conf/lpnmr/SyrjanenN01} for details). 

Let us recall the syntax of weight rules via an example. 
Consider the following rule $r_1$:
\begin{lstlisting}[style=asp-style]
less_eq $\derives$ 2 $\leq$ #sum{1:a(1); 3:a(3)}.
\end{lstlisting}
In the numeric format, each atom is associated to a positive integer identifier.
Suppose that \textit{a(1)} is mapped to $1$, \textit{a(3)} to $2$ and \textit{less\_eq} to $3$.
According to the numeric format, $r_1$ would be converted as:
\begin{lstlisting}[style=asp-style-inline,keywordstyle=\linespread{1}\ttfamily,
	basicstyle=\linespread{1}\ttfamily
 ]
$l_1:$ 1 3 1 0 4 
$l_2:$ 5 4 2 2 0 1 2 1 3 
\end{lstlisting}
Line $l_1$ is a basic rule in the numeric format jargon, stating that the atom with id $3$ is true iff the atom with id $4$ is true as well.
The atom with id $4$ is defined via the weight rule in line $l_2$. 
Weight rules start with a $5$ and have the following form: 
\begin{lstlisting}[style=asp-style-inline,keywordstyle=\linespread{1}\ttfamily,
	basicstyle=\linespread{1}\ttfamily
]
5 agg bound #lits #negative negative positive weights
\end{lstlisting}
Each literal within the aggregate has associated a weight: $1$ is the weight of \textit{a(1)} and $3$ is the weight of \textit{a(3)}.
Intuitively, the atom \textit{less\_eq} (id $3$) is true iff the bound, i.e., $2$ is less or equal than the sum of the weights of true literals.

In the proposed extension the bound and all weights are rationals, thus a factor equal to the $lcm$ of all their denominators is multiplied to each of them. 
In this way, we get only rationals with denominator equal to $1$, representable as integers, maintaining proportionality.
For instance, consider the rule $r_2$: 
\begin{lstlisting}[style=asp-style]
less_eq $\derives$ 2 $\leq$ #sum{3/4:a(3/4); 3:a(3)}.
\end{lstlisting}
Assuming that \textit{a(3/4)} is mapped to the id $1$, \textit{a(3)} to $2$ and \textit{less\_eq} to $3$, $r_2$ would be converted in the numeric format as:
\begin{lstlisting}[style=asp-style-inline,keywordstyle=\linespread{1}\ttfamily,
	basicstyle=\linespread{1}\ttfamily
]
1 3 1 0 4 
5 4 8 2 0 1 2 3 12 
\end{lstlisting}


Similarly, in minimize rules used to represent weak constraints weights are \rationals. For each level $l$, a factor equal to the $lcm$ of all the denominators of weights at level $l$ has to be multiplied to each of these weights.



\section{Handling Non-Integer Domains with ASP}\label{sec:motivations}
Typically, the inability of ASP to handle non-integer arithmetic is dammed using workarounds that delegate the manipulation of non-integers to the user while remaining invisible to ASP systems. This means that the presence of non-integers is managed by the user themselves. For instance, hybrid reasoning is utilized in some ASP-based robotics applications\cit{DBLP:journals/aim/ErdemGL16,DBLP:journals/ki/ErdemP18}.


To illustrate more commonly adopted workarounds, let us consider as example, a real world scenario taken from CityBench\cit{DBLP:conf/semweb/AliGM15}, a Smart City benchmark;
it consists of $13$ continuous queries that require to reason on dynamic data streams, coming from sensors scattered throughout the city of Aarhus in Denmark.
We consider next the third query, $Q_3$: given a planned journey, it requires to compute the average congestion level and the estimated travel time to a destination.
Focusing on the computation of the average congestion level, let us assume that input facts over the predicates \textit{journey} and \textit{roadLength} define the roads involved in the journey and the lengths of each road, respectively.
The following is an ASP modelling for $Q_3$:
\begin{lstlisting}[style=asp-style]
$r_1:$ congestionLevel(Road, CL) $\derives$ journey(Road), 
$\hspace{2.5em}$ vehicleCount(Road, C), roadLength(Road, L), CL = C / L.
$r_2:$ totCongestionLevel(T) $\derives$ 
$\hspace{2.5em}$ #sum{CL, Road: congestionLevel(Road, CL)} = T.
$r_3:$ roadsCount(N) $\derives$ #count{Road: journey(Road)} = N.
$r_4:$ avgCongestionLevel(Avg) $\derives$ totCongestionLevel(S), numRoads(X), 
$\hspace{2.5em}$ Avg = S / X.
\end{lstlisting}

Rule $r_1$ is used to compute the congestion level for each road of the planned journey: fixed a road, it is the ratio between the number of vehicles and the length of the road (in meters), i.e., the number of vehicles per each meter. 
Rule $r_2$ sums up the so computed congestion levels determining the congestion level of the whole journey.
Rule $r_3$ counts the number of roads of the journey. 
Rule $r_4$ determines the average congestion level of the journey as the ratio between the total congestion level and the number of roads.
Clearly, congestion levels should be computed as \noninteger numbers, as their values influence the computation of the overall congestion level and the loss of accuracy is accumulated up to the query answer. 
When no support is available for \noninteger numbers, a possible workaround could be to rely on integer constants and, whenever a division occurs, in order to keep track of decimal digits that would be truncated, one can multiply the dividend by a power of 10, say $10^p$.
This trick allows to maintain the most significant $p$ decimal positions produced by the ratio. 
A final post-processing is then needed to convert back the result and compute the average congestion level as \noninteger numeric value.
In the case of the program above, if we want, for instance, to keep $2$ decimal digits, $r_1$ and $r_4$ could be rewritten as follows:
\begin{lstlisting}[style=asp-style]
$r_1':$ congestionLevel(R, CL) $\derives$ journey(R), vehicleCount(R, C), 
$\hspace{2.5em}$ roadLength(R, L), C_INT = C * 100, CL = C_INT / L.
$r_4':$ avgCongestionLevel(Avg) $\derives$ totCongestionLevel(S), 
$\hspace{2.5em}$numRoads(X), Avg_INT = S * 100, Avg = Avg_INT / X.
\end{lstlisting}
Consider now a journey passing through three roads $x$, $y$ and $z$ whose lengths are $1000$, $1500$ and $3000$ meters and the vehicle counts are $30$, $55$ and $80$, respectively.
The congestion level of road $x$ is $3000/1000 = 3$; for the road $y$, $5500/1500$ and for road $z$, $8000/3000$ that mainstream ASP systems would compute as integer divisions as $3$ and $2$, respectively. 
The total congestion level is $8$. 
The average is computed as $266$ (i.e., $800/3$).
At this point, given that two multiplications by $100$ occurred, the outcome has to be divided two times by $100$ during a post-processing phase, obtaining that the final average is $0.0266$ and only the two most significant digits are actually reliable, thus the average is estimated as $0.02$.
The proposed semantics would instead yield an exact result of $\frac{7}{225} = 0.03\overline{1}$, which matches the actual average. This happens when in the input program \rational terms are in the format $(1)$ or $(2)$, as no approximations are necessary, and all computed results are precise thanks to the closure property of \rational numbers under all four fundamental mathematical operations.
In addition to its lack of precision, it is worth to note that the workaround is considerably less declarative. 
In general, each multiplication must be considered to ensure that the output is appropriately adjusted. 
Additionally, when input facts contain non-integer numbers, a pre-processing step is necessary, in which these numbers are converted to integers by multiplying them by $10^p$ and potentially rounding them.
In domains that demand high reactivity, such pre/post processing may become a bottleneck.

A different workaround consists instead in using string terms in place of \noninteger numbers. 
This requires to $i)$ quote all such numbers and $ii)$ in this light, properly manage arithmetic operations and comparisons given that, when applied on strings, their default behaviour may not be as expected (e.g., \quo{10.1}$<$\quo{2.1}).
A solution is to rely on hybrid approaches, based on linguistic constructs not part of ASP-Core-2, allowing to change such default behaviour defining a custom one.
In this respect, there are still no standards and each system provides its own functionalities. 
Some ASP systems offer the possibility of specifying within rules external arbitrary functions evaluated at grounding time and defined via programming languages.
For instance, \clingo\cit{DBLP:journals/tplp/GebserKKS19} provides an integration with the Lua and Python scripting languages; the language of \dlvdue\cit{DBLP:conf/lpnmr/AlvianoCDFLPRVZ17} includes external literals, whose semantics can be externally defined via Python. 
Nonetheless, the usage of such features can result in a less declarative and more intricate modelling;
e.g., in the case of \dlvdue, $Q_3$ could be modelled as follows.
\begin{lstlisting}[style=asp-style]
$r''_1:$ congestionLevel(R,CL) $\derives$ journey(R), vehicleCount(R, C), 
$\hspace{2.5em}$ roadLength(R, L), &div(C, L; CL).
$aux_1:$ precedes(R1, R2) $\derives$ journey(R1), journey(R2), R1 < R2.
$aux_2:$ successor(X, Y) $\derives$ precedes(X, Y), not inBetween(X, Y).
$aux_3:$ inBetween(X, Y) $\derives$ precedes(X, Z), precedes(Z, Y).
$aux_4:$ first(X) $\derives$ journey(X), not hasPredecessor(X).
$aux_5:$ last(X) $\derives$ journey(X), not hasSuccessor(X).
$aux_6:$ hasPredecessor(X) $\derives$ successor(Y, X).
$aux_7:$ hasSuccessor(Y) $\derives$ successor(Y, X).
$aux_8:$ partialSum(R, CL) $\derives$ first(R), congestionLevel(R, CL).
$aux_9:$ partialSum(R2, S) $\derives$ successor(R1, R2), partialSum(R1, PS), 
$\hspace{2.5em}$ congestionLevel(R2, CL), &sum(PS, CL; S).
$r''_2:$ totCongestionLevel(T) $\derives$ last(R), partialSum(R, T).
$r_3:$ roadsCount(N) $\derives$ #count{R: journey(R)}=N.
$r''_4:$ avgCongestionLevel(Avg) $\derives$ numRoads(X), 
$\hspace{2.5em}$ totCongestionLevel(S), &div(S, X; Avg).
\end{lstlisting}
Rule $r_1$ is replaced by $r''_1$ demanding to the external atom \textit{\&}\textbf{div}\textit{(C,L;CL)} the division \textit{CL = C / L}. 
To obtain a more accurate result, the semantics could be defined via the following Python function that converts back strings into \float numbers according to the Python representation (IEEE 754), computes the division and returns the result as string: 
\begin{lstlisting}[style=python3]
def div(A,B):
    return str(float(A)/float(B))
\end{lstlisting}
Rule $r_2$ is re-adapted as well and, in order to compute each partial sum as a \float number, auxiliary rules $aux_1 \dots aux_9$ are needed; \textit{\&{\bf sum}(PS,CL;S)} in rule $aux_9$ corresponds to $S=PS+CL$ and it is implemented as:
\begin{lstlisting}[style=python3]
def sum(A,B): 
    return str(float(A)+float(B)).
\end{lstlisting}
Rule $r_4$ is also updated, becoming $r''_4$, to compute \float divisions again via \&{\bf div}.
This workaround permits to achieve a higher accuracy. 
However, the encoding could be less intuitive, pre/post processing steps are needed and the frequency of external calls may degrade time performance.
Moreover, the declarative purpose of ASP is \quo{compromised} by the need for imperative code defining parts of the semantics.
More importantly, this workaround may lead to \quo{faulty} results as in the case of the following program $P_1$:
\begin{lstlisting}[style=asp-style]
a("1.0").   a("0.0000000000000001"). 
result(W) $\derives$ a(X), a(Y), a(Z), &sum(X, Y, Z; W).
\end{lstlisting}
where the semantics of the external atom \textit{\&{\bf sum}(X,Y,Z;W)} is defined according to the following Python code:
\begin{lstlisting}[style=python3]
def sum(A,B,C): 
    return str(float(A)+float(B)+float(C))
\end{lstlisting}
The rule consists in a Cartesian product of all possible ground atoms over the predicate $a/1$. 
Because the addition of floating point numbers is not associative, the sum of the same three values can be different.
In particular, the result of the sum is:
\begin{itemize}
    \item $1.0000000000000002$ if: \\
    $\{X\mapsto \text{\em{"0.0000000000000001"}}$, 
    $Y\mapsto \text{\em{"0.0000000000000001"}}$, 
    $Z\mapsto \text{\em{"1.0"}}\}$
    \item $1.0$ if: \\
    $\{X\mapsto \text{\em{"1.0"}}, 
    Y\mapsto \text{\em{"0.0000000000000001"}}, 
    Z\mapsto \text{\em{"0.0000000000000001"}}\}$.
\end{itemize}
Preventing such issues is in charge of the user: the Python code must ensure the reproducibility of the sum by utilizing appropriate methods\cit{DBLP:conf/arith/DemmelN13} or relying on rationals rather than on floating-point representations. 
It is evident that these workarounds may deter less experienced users, since a deep understanding about how the system evaluates the input program is required.

A similar solution can be based on \clingo Python\footnote{https://potassco.org/clingo/python-api/5.6} or C API\footnote{https://potassco.org/clingo/c-api/5.6}, which provides a rich set of functions to manage, ground and solve logic programs. For instance, the program $P_1$ can be modelled via \clingo Python API as illustrated below.

\begin{lstlisting}[style=python3, morekeywords={script,end}]
#script (python)
import clingo
class Context:
    def sum(context,x,y,z):
        w = float(x.string) + float(y.string) 
            + float(z.string)
        return clingo.String(str(w))
def main(prg):
    prg.ground([("base", [])], context=Context())
    prg.solve()
#end.
\end{lstlisting}
\begin{lstlisting}[style=asp-style]
a("1.0").    a("0.0000000000000001").
result(@sum(X,Y,Z)) $\derives$ a(X), a(Y), a(Z). 
\end{lstlisting}
It is important to note that, compared to external atoms, this solution is even more powerful and efficient as it allows the ASP developer to even control the grounding and solving process\cit{DBLP:journals/tplp/KaminskiRSW23}.
However, the same concerns about potentially counter-intuitive results over floating-points, the need for advanced knowledge, and the resort to imperative programming apply in this case as well. 

\section{Implementation}\label{sec:implementation}
Along with the extended ASP language, we provide in \system an implementation thereof.
\system is an ASP instantiator (or grounder), is available at \url{https://demacs-unical.github.io/I-DLV}.
\system is compatible with state-of-the-art ASP solvers~\cite{DBLP:journals/ia/CalimeriFPZ17} as it implements the aforementioned intermediate numeric output format (see Section\re{sec:technical}). 

For \rational terms of form $(3)$, in \system the maximum number of decimal digits $f$ that is kept, is by default set to $6$.
Moreover, \system provides a command-line option allowing the user to specify a desired $f$ value up to $6$.
Furthermore, an other option permits to specify how \rational terms appearing in the answer sets have to be printed. 
In particular, a \rational not reducible to an integer can be printed as fraction (i.e., form $(1)$) or be outputted as the value obtained approximating the division it represents (i.e., form $(3)$).
In this latter case, \system adopts the same rounding policy used for input \rationals of form $(3)$, as described in Section\re{sec:technical}.

The implementation has been endowed with a set of well-defined mathematical functions, commonly of use when dealing with \rational numbers. 
The syntax of such functions is inspired by \textit{dlvhex}\cit{DBLP:journals/tplp/Redl16,DBLP:conf/aiia/CalimeriFPZ17}\footnote{\url{https://github.com/DeMaCS-UNICAL/I-DLV/wiki/External-Computations,-Interoperability-and-Linguistic-Extension}}; they correspond to possibly negated literals of form $\&fun(i_1,\dots,i_n ; o_1,\dots,o_m)$, where $fun$ is an identifier recalling the corresponding function, $i_1,\dots,i_n$ and $o_1,\dots,o_m$ are input and output terms, respectively.
The set of supported functions is reported in Table~\ref{tab:mathematical-builtins}.
For instance, \{\textit{a(3/4), pow(9/16)}\} is the only answer set of the program:
\lstinline[style=asp-style-inline]
|a(3/4). pow(X,Y) $\derives$ a(X), &pow(X,2;Y).|

\renewcommand{\arraystretch}{1.3}
\begin{table}
\centering
\caption{The set of functions for \rational terms.}
\label{tab:mathematical-builtins}
\begin{tabular}{l|l} 
\hline
\multicolumn{1}{c|}{\textbf{Function}} & \multicolumn{1}{c}{\textbf{Semantics}}  \\ 
\hline
\textbf{truncate(X;Z)}   & Assigns to $Z$ the value of $X$ truncated as integer\\
\hline
\textbf{round(X;Z)}      & \begin{tabular}[c]{@{}l@{}}Assigns to $Z$ the value of $X$ rounded to the\\nearest integer\end{tabular}\\
\hline
\textbf{ceil(X;Z)}       & \begin{tabular}[c]{@{}l@{}}Assigns to $Z$ the smallest integer value that is\\not less than $X$\end{tabular}\\
\hline
\textbf{floor(X;Z)}      & \begin{tabular}[c]{@{}l@{}}Assigns to $Z$ the value of $X$ rounded downward\\as integer\end{tabular}\\
\hline
\textbf{pow(X,E;Z)}      & \begin{tabular}[c]{@{}l@{}}Assigns to $Z$ the value of $X$ to the power of $E$,\\i.e., $X^{E}$\end{tabular}\\
\hline
\textbf{abs(X;Z)}        & Assigns to $Z$ the absolute value of $X$\\
\hline
\end{tabular}
\end{table}


\section{Conclusions}\label{sec:conclusions}
This work puts forth a proposal for broadening the scope of ASP in the AI field, by incorporating \rational terms and enabling the use of numbers beyond integers in a purely declarative manner.
This proposal enhances the practical applicability of ASP in various contexts and bridges the gap with other logical formalisms, which are already adept at handling \noninteger domains. Additionally, we provide a concrete contribution by presenting an implementation that adheres to the \quo{ground\&solve} approach.
To the best of our knowledge, the extension of ASP in this direction has never been explored, as evinced by the fact that no ASP system allows for non-integer numeric types.
\bibliographystyle{plain}
\bibliography{references}
\end{document}